\pdfoutput=1

\documentclass[11pt]{article}

\usepackage[preprint]{acl}

\usepackage{times}
\usepackage{latexsym}
\usepackage{amsmath}
\usepackage{adjustbox}
\usepackage{longtable}

\usepackage{amssymb}
\usepackage{bm}
\usepackage{bbm}
\usepackage{cleveref}
\usepackage{makecell}
\usepackage{xcolor}
\usepackage[T1]{fontenc}

\usepackage[utf8]{inputenc}

\usepackage{microtype}

\usepackage{inconsolata}

\usepackage{graphicx}
\usepackage{adjustbox}
\usepackage{colortbl}
\usepackage{overpic}

\title{Alignment for Efficient Tool Calling of Large Language Models}

\author{Hongshen Xu\textsuperscript{1}\footnotemark[1], Zihan Wang\textsuperscript{1}\footnotemark[1], Zichen Zhu\textsuperscript{1}, Lei Pan\textsuperscript{2},\\ \textbf{Xingyu Chen}\textsuperscript{1}, \textbf{Lu Chen\textsuperscript{1}, Kai Yu\textsuperscript{1}\footnotemark[2]}\\
  \textsuperscript{1}X-LANCE Lab, Department of Computer Science and Engineering \\
  MoE Key Lab of Artificial Intelligence, AI Institute\\
  Shanghai Jiao Tong University, Shanghai, China 
  \\
 \textsuperscript{2}AISpeech Co., Ltd., Suzhou, China\\
  \texttt{\{xuhongshen, kai.yu\}@sjtu.edu.cn} \\}

\begin{document}
\maketitle
\begin{abstract}
Recent advancements in tool learning have enabled large language models (LLMs) to integrate external tools, enhancing their task performance by expanding their knowledge boundaries. However, relying on tools often introduces trade-offs between performance, speed, and cost, with LLMs sometimes exhibiting overreliance and overconfidence in tool usage. This paper addresses the challenge of aligning LLMs with their knowledge boundaries to make more intelligent decisions about tool invocation. We propose a multi-objective alignment framework that combines probabilistic knowledge boundary estimation with dynamic decision-making, allowing LLMs to better assess when to invoke tools based on their confidence. Our framework includes two methods for knowledge boundary estimation—consistency-based and absolute estimation—and two training strategies for integrating these estimates into the model’s decision-making process. Experimental results on various tool invocation scenarios demonstrate the effectiveness of our framework, showing significant improvements in tool efficiency by reducing unnecessary tool usage.
\end{abstract}

\renewcommand{\thefootnote}{\fnsymbol{footnote}}
\footnotetext[1]{Equal contributions.}
\footnotetext[2]{The corresponding author is Kai Yu.}

\section{Introduction}

\begin{figure}
    \centering

    \begin{overpic}[width=\linewidth]{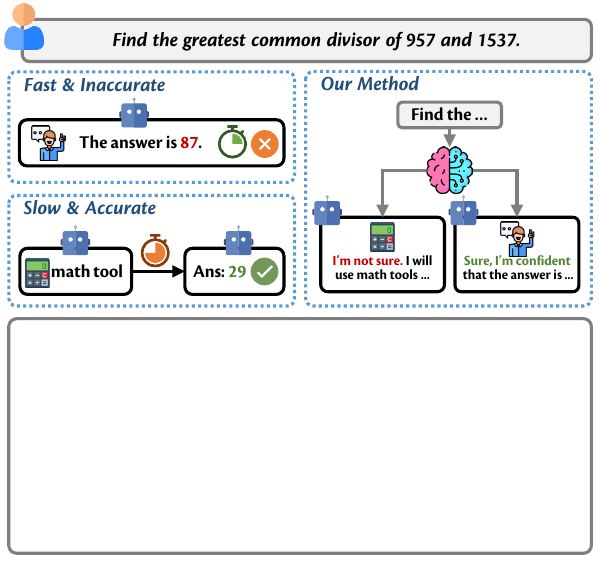}
        \put(2.5,2.5){\includegraphics[width=0.95\linewidth]{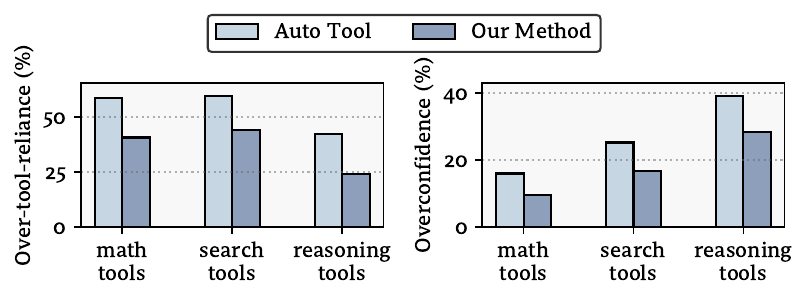}}
    \end{overpic}

    \caption{Our method effectively enables LLMs to switch between answering independently and calling tools (upper part), thereby reducing the model's over-reliance and overconfidence in tools (lower part).}
    \label{fig:progress}
    \vspace{-5mm}
\end{figure}

The objective of tool learning is to enable large language models (LLMs; \citealp{geminiteam2023gemini,achiam2023gpt,dubey2024llama}) to acquire the capability to effectively utilize external tools, thereby enhancing their performance across various downstream tasks~\cite{schick2023toolformer,hao2023toolkengpt, hsieh2023tool, tang2023toolalpaca}. Tools can be regarded as extensions of an LLM's knowledge or capability boundaries. By invoking tools, models can accomplish tasks beyond their knowledge boundaries and even access information from different modalities~\cite{zeng2022socratic}.

While tools can enhance LLM's task performance, it is important to note that solving tasks through tool invocation often requires more steps, longer completion times, and additional tool-calling costs. For example, in question-answering scenarios involving search tools, the model must first generate a query for the retrieval tool, wait for the search results, and then process these results to produce a final answer. In contrast, direct answering involves simply generating a response. This introduces a trade-off problem between performance and speed. Unfortunately, recent studies have shown that O1-like LLMs struggle to strike a balance between these two aspects: exhibit \textit{overthinking}~\cite{chen2024not} in simple reasoning tasks and \textit{underthinking}~\cite{wang2025thoughts} in more difficult ones.  Similarly, we observe that the same issue arises in tool usage scenarios. Current LLMs exhibit \textbf{\textit{over-tool-reliance}}, invoking tools even when tasks could be completed independently, while also exhibiting \textbf{\textit{overconfidence}} by refusing to use tools when necessary. This inconsistency mirrors the challenges faced by O1-like models, undermining the model's tool intelligence and increasing task completion costs in real-world scenarios.

In this work, we aim to improve how LLMs decide when and how to use external tools for task completion. The main challenge is aligning the model’s behavior with its knowledge boundaries, allowing it to determine when a tool is needed based on its confidence. Instead of treating the model’s knowledge as simply "known" or "unknown"~\cite{yang2023alignment}, we propose a more nuanced approach that accounts for uncertainty. This approach recognizes an "\textbf{\textit{uncertain region}}" where the model assigns probabilistic estimates to its knowledge, enabling better decision-making that balances task success and tool usage costs.

We introduce an alignment framework for efficient tool calling that combines probabilistic knowledge boundary estimation with dynamic decision-making. Our approach has two main components: 1) \textbf{Knowledge Boundary Estimation}: we propose two methods to assess the model’s knowledge: consistency-based estimation based on agreement and using external ground truth to evaluate the average accuracy of multiple model samplings. 2) \textbf{Knowledge Boundary Modeling}: we construct different data to exhibit \textit{implicit modeling}, where the model makes decisions based on predefined thresholds of knowledge certainty, and \textit{explicit modeling}, where the model outputs both an answer and a confidence score. This framework helps the model use tools more efficiently, invoking them only when necessary, thus improving performance while reducing costs. Our approach is evaluated across multiple tool-use scenarios, demonstrating a significant reduction in unnecessary tool invocation and an improvement in overall tool efficiency. Our contributions can be summarized as follows:
\begin{itemize}
    \item We propose a multi-objective alignment framework for efficient tool invocation, along with corresponding evaluation metrics. 
    \item We propose the tool alignment algorithms and corresponding data generation methods.
    \item We conduct extensive experiments across multiple tool invocation scenarios, demonstrating the effectiveness of our approach.
\end{itemize}

\section{Related Work}

\subsection{LLM Alignment}
LLM alignment seeks to train language models to act in accordance with the user’s intent, utilizing methods such as supervised fine-tuning \citep{wei2021finetuned, chung2022scaling, zhang2023instruction}, direct preference optimization (DPO) \citep{rafailov2024direct}, or reinforcement learning from human feedback (RLHF) \citep{stiennon2020learning, ouyang2022training, glaese2022improving}. Most works focus on enhancing the instruction-following capabilities \citep{sanh2021multitask, wei2021finetuned}, helpfulness \citep{ding2023enhancing, xu2023wizardlm}, harmlessness \citep{solaiman2021process, bender2021dangers}, and honesty \citep{cui2023ultrafeedback, yang2023improving} of LLMs. In addition, some works proposed aligning models with their knowledge boundaries~\cite{xu2024rejection, yang2023alignment, zheng2025enhancing}, specifically by training LLMs to reject unknown questions. However, these approaches assume a binary view of the model’s knowledge boundary—either the model knows the answer or it does not. In contrast, our work posits that knowledge boundaries are more nuanced and exist within a gray area. We propose dynamically determining the model’s behavior within this ambiguous region, depending on the specific application scenario.

\subsection{Tool Learning}
Recent advancements in tool learning have enabled LLMs to effectively integrate external tools, enhancing real-time knowledge retrieval, multimodal functionalities, and domain-specific expertise ~\citep{yang2023chatgpt, gupta2023visual, jin2023genegpt}. Methods range from leveraging in-context learning for tool descriptions and demonstrations~\citep{hsieh2023tool} to explicit training on tool-enriched datasets~\citep{patil2023gorilla, tang2023toolalpaca, qin2023toolllm}. Some works have also investigated how to accomplish tasks within a limited number of tool invocations~\cite{zheng2024budget} and how to call tools more reliably~\cite{xu2024reducing, gui2024look}. However, previous research on tool invocation has largely overlooked the correlation between tool usage and the model's knowledge boundaries. Additionally, there has been no unified evaluation metric proposed for assessing efficient tool invocation. 

\section{Problem Formulation}

\subsection{LLM Alignment}

With the rapid development of large language models (LLMs), ensuring their alignment with human instructions, preferences, and values has become a crucial research area~\citep{wang2024essence}. Alignment approaches are designed to optimize model responses based on predefined objectives such as helpfulness, truthfulness, and safety. Specifically, given an input prompt $x_i$ and an alignment goal \textit{helpfulness}, we employ the following scoring principle to represent the alignment objective:
\begin{align}
s(x, y_h) > s(x, y_u),
\end{align}
where $y_h$ and $y_u$ represent a helpful response and an unhelpful response, respectively. The preference order can be determined through human annotation~\citep{ouyang2022training} or a scoring model~\citep{gao2023scaling} trained with human preference data. The collected preference data can be further leveraged to train reward models or fine-tune LLM policies, thereby improving alignment with human expectations.


\subsection{Multi-Objective Alignment for Efficient Tool Calling}

While alignment with helpfulness is essential, efficient tool calling introduces additional alignment challenges. A well-aligned LLM should not only provide helpful responses but also minimize unnecessary tool usage, as excessive tool calls increase inference latency and computational costs. Therefore, we propose a multi-objective alignment framework that balances \textit{helpfulness} and \textit{tool cost}.

First, we define alignment objectives separately for helpfulness and tool cost. The helpfulness alignment objective follows:
\begin{align}
s(x, y_c) > s(x, y_w),
\end{align}
where $y_c$ represents a correct response, and $y_w$ represents an incorrect response. Simultaneously, for tool cost, we define:
\begin{align}
s(x, y_n) > s(x, y_t),
\end{align}
where $y_n$ represents a response without tool usage, and $y_t$ represents a response with tool usage. Combining these two objectives, our final alignment formulation becomes:
\begin{align}
s(x, y_{nc}) > s(x, y_{tc}) > s(x, y_{nw}) > s(x, y_{tw}),
\end{align}
where $y_{nc}$, $y_{tc}$, $y_{nw}$, $y_{tw}$ represent correct responses without tool usage, correct responses with tool usage, incorrect responses without tool usage, and incorrect responses with tool usage, respectively.
This ordering reflects the principle that an ideal LLM should solve problems independently whenever possible, resorting to tool usage only when necessary, while also avoiding incorrect answers and unnecessary tool calls.

\subsection{Evaluation Methodology for Efficient Tool Calling}

To quantify the tradeoff between helpfulness and tool cost, we define a \textbf{benefit-cost utility} function as follows:
\begin{align}
u(y) = \mathbbm{1}_{helpfulness}(y)  - \alpha \cdot \mathbbm{1}_{cost}(y),
\end{align}
where $\mathbbm{1}_{helpfulness}(y)$, $\mathbbm{1}_{tool}(y)$ equal to 1 when the response $y$ is correct or contains tool calling, respectively. $\alpha$ represents the cost associated with tool usage. The overall utility of a model on a dataset with $N$ samples is then computed as:
\begin{align}
\text{Utility} &= \frac{1}{N} 
\sum_{i=1}^{N} u(y_i) = \text{Acc} - \alpha\cdot \text{TR},
\end{align}
where Acc and TR represent the overall accuracy and tool usage ratio on the dataset, respectively.

The parameter $\alpha$ is crucial, as it determines the relative penalty of tool usage. A larger $\alpha$ indicates a higher sensitivity to cost or a greater penalty for invoking tools. If $\alpha$ is too high, the model may completely avoid tool usage, even when necessary. Conversely, if $\alpha$ is too low, the model may overuse tools. Therefore, selecting a moderate $\alpha$ ensures a balanced tradeoff between efficiency and effectiveness. Furthermore, the cost of tool usage varies across different tasks and tools. To account for these differences, $\alpha$ can be set dynamically based on the specific tool being used. Empirically, in our study, we assign $\alpha$ values of 0.2, 0.4, and 0.6 to calculators, search engines, and external LLM reasoning, respectively. The different $\alpha$ values reflect the increasing computational cost and inference latency associated with these tools.

\section{Methodology}

\begin{figure*}[t!]
    \centering
    \includegraphics[width=\linewidth,trim=110 90 100 120,clip]{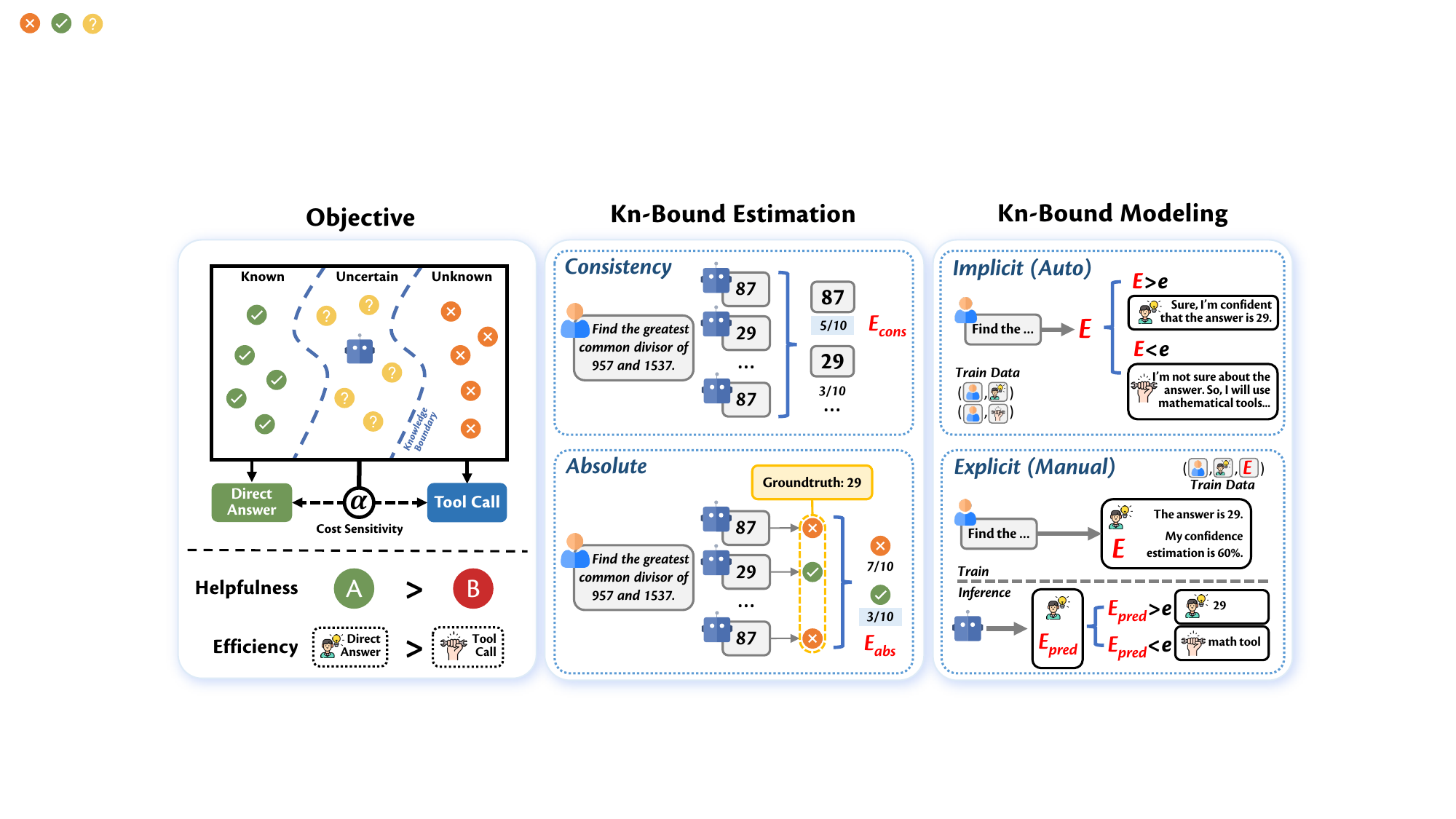}
    \caption{The overall pipeline of knowledge boundary modeling methods.} 
    \label{fig:overview}
\end{figure*}

\subsection{Framework for Efficient Tool Learning}
The key to enabling efficient tool calling lies in aligning LLMs with their own knowledge boundaries. Unlike a binary classification of knowledge into "known" and "unknown," human cognition—and by extension, LLMs—operates within a spectrum. As shown in the left part of Figure~\ref{fig:overview}, there exists a large "\textit{\textbf{uncertain region}}" where the model can only assign a probabilistic estimate to its knowledge. Previous works that enforce a strict binary classification fail to capture this nuanced understanding, leading to inaccurate estimations and suboptimal tool invocation strategies.

To achieve effective tool use, the model must first develop an awareness of its knowledge boundaries and then leverage this understanding to adjust its decision-making process. This perspective aligns with the efficiency objective discussed in prior sections: a model that perceives knowledge in binary terms will struggle to adjust its behavior under varying cost considerations (represented by $\alpha$). If a model simply categorizes knowledge as either "known" or "unknown," it will either always invoke a tool for uncertain cases or always answer directly, ignoring cost-sensitive optimization.

We propose a solution where the model learns to estimate its knowledge uncertainty probabilistically rather than making binary classifications. This allows for greater flexibility in tool invocation. Depending on different values of $\alpha$ (which represent different real-world tool costs), we can train the model to dynamically adjust its behavior. This can be implemented implicitly through controlled training data distributions or explicitly by having the model output confidence estimates that can be thresholded at inference time to determine whether a tool should be invoked.

\subsection{Estimating Knowledge Boundaries}

We propose two methods for knowledge boundary estimation as shown in the middle part of Figure~\ref{fig:overview}:

\paragraph{Consistency-Based Estimation} This method relies on self-consistency. We assume that if a model produces highly consistent outputs across multiple samples for a given question, it possesses a stronger grasp of the underlying knowledge. To operationalize this, we measure the variance in the model’s sampled responses and use it as an indicator of knowledge certainty. Higher consistency implies greater confidence in the model’s knowledge.

\paragraph{Absolute Estimation via Ground Truth} While consistency-based estimation is useful, it does not directly leverage external validation. To address this, we introduce an absolute estimation method based on ground truth correctness. We repeatedly sample model responses for the same question and compute the average accuracy using ground truth. This provides an externally validated measure of the model’s knowledge, correcting for potential biases in self-estimation.

\subsection{Training Approaches}

To integrate knowledge boundary estimation into the model’s behavior, we employ two SFT strategies as shown in the right part of Figure~\ref{fig:overview}: implicit modeling and explicit modeling.

\paragraph{Implicit Modeling} In this approach, the model is trained to directly output actions (either answering directly or invoking a tool) based on pre-defined decision rules. Specifically, we sort all training samples based on their estimated knowledge scores and set a threshold: samples above this threshold are labeled for direct answering, while those below it are labeled for tool invocation. Since different values of $\alpha$ correspond to different tool usage preferences, we train separate SFT models with varying thresholds to adapt to different scenarios. This method is efficient during inference, as the model only needs to generate a single response per query. However, it requires multiple rounds of training for different values of $\alpha$.

\paragraph{Explicit Modeling} Unlike implicit modeling, explicit modeling trains the model to output both an answer and an associated knowledge confidence score. This allows dynamic adjustment of tool invocation decisions at inference time without requiring separate SFT models for different $\alpha$ values. During inference, we set a threshold on the confidence score: if the score is above the threshold, the model answers directly; otherwise, it invokes a tool. This approach eliminates the need for retraining but introduces additional inference latency, as each query requires both an answer and an uncertainty estimation before deciding whether to use a tool.

Each method has its advantages and drawbacks.
Implicit Modeling has Faster inference (single response generation) but requires multiple training runs for different $\alpha$ values. Explicit Modeling is more flexible at inference time (threshold tuning without retraining) but slower due to the two-step generation process. In our experiments, we evaluate both approaches to determine the most effective strategy for efficient tool calling.
\section{Experiments}
\subsection{Experiment Setup}
\subsubsection{Task Scenarios}

We evaluate our approach across three scenarios, each requiring a specific external tool: symbolic computation via a calculator, factual retrieval using a retrieval-augmented generation \cite{gao2023retrieval} system, and complex reasoning with a strong reasoning model.(see Appendix~\ref{sec:setup} for more detailed experimental setup.).

\noindent\textbf{Arithmetic Computation (Calculator).} 
To assess numerical reasoning capabilities, we construct an arithmetic dataset following ~\citet{liu2023goat}. Input numbers are sampled from a logarithmic scale to ensure diverse magnitudes, with minimal duplication probability. We incorporate hundreds of instruction templates generated by ChatGPT to increase linguistic diversity. Computation is performed using a symbolic calculator, implemented via code execution for mathematical evaluation.

\noindent\textbf{Knowledge-based QA (Retrieval-Augmented Generation).} 
To evaluate factual knowledge retrieval, we use TriviaQA~\citep{joshi2017triviaqa}, a widely used question-answering dataset. We sample 10,000 instances for training and use the 11,313 instance development set for evaluation, as the official test set ground truth is unavailable. To enhance factual accuracy, we integrate a retrieval system, leveraging Pyserini~\citep{lin2021pyserini}—a Python toolkit designed for reproducible information retrieval with sparse and dense representations. 

\noindent\textbf{Complex Reasoning (Reasoning Model).} 
For evaluating multi-step reasoning tasks, we employ the MATH dataset~\citep{hendrycksmath2021}. We use the original train-test split. Given the inherent complexity of mathematical reasoning, we utilize DeepSeek-R1~\cite{deepseekai2025deepseekr1incentivizingreasoningcapability}which provides strong reasoning capabilities. However, it comes with a trade-off: higher computational cost and slower inference speed.

\subsubsection{Baselines}
The baseline methods are categorized into two major groups: \textit{Prompt-based} and \textit{Uncertainty-based}. All prompts used  are listed in Appendix~\ref{sec:Prompts}.
\paragraph{Prompt-based}  
Prompt-based methods govern how the model interacts with external tools and determines its tool usage behavior. The Baseline (w/o tool) approach has the model answer queries entirely on its own, relying only on internal knowledge. The Baseline (all tool) forces the model to always invoke a tool. The Auto tool method allows the model to decide when to use a tool based on its estimated confidence. ICL tool (10-shot) provides the model with 10 example interactions (5 correct, 5 incorrect) to better guide its decision on whether to answer directly or use a tool.


\paragraph{Uncertainty-based.}  
Uncertainty-based methods estimate the confidence of model-generated answers, which we leverage to determine the optimal utility by searching for the best confidence threshold. We explore four approaches: Raw logits~\cite{lyu2024calibrating}, P(True)~\cite{kadavath2022language}, Verbalized Confidence~\cite{tian2023just}, and Agreement (Self-Consistency)~\cite{lyu2024calibrating}, each providing a different way to assess model confidence (see Appendix~\ref{sec:uncertainty} for details).

\subsubsection{Training Details}
We use two baseline models: \textsc{LLama-3.1-8B-Instruct} and \textsc{Qwen-2.5-7B-Instruct}.To align with our experimental setup, we customize the DeepSpeed-Chat~\cite{yao2023deepspeed} framework. The training process adopts a learning rate of \(5 \times 10^{-5}\) and a batch size of 128. All other training parameters are set to the default parameters in DeepSpeed-Chat.By default, 10,000 samples are used for Supervised Fine-Tuning. All models undergo training for 2 epochs on A800 GPUs.

\subsection{Main Results}

Table~\ref{tab:main_results} compares the performance of all evaluated methods. Our approach achieves the highest \textit{utility} scores across three datasets, demonstrating its effectiveness in balancing task success and tool efficiency.
Among our methods, Absolute-based knowledge boundary estimation outperforms Consistency-based estimation, as external supervision via ground truth labels enables more accurate boundary estimation and better tool invocation decisions.
Our approach maintains accuracy comparable to the best models while reducing tool usage by nearly 50\% compared to fully automatic baselines. It also matches the Baseline (All Tools) in accuracy while significantly lowering reliance on external tools, reducing computational costs.
Our training-based method further enhances efficiency compared to Auto Tool, achieving better performance while reducing tool usage. This validates the effectiveness of refining tool invocation alignment with the model's internal knowledge boundary.

\begin{table*}[!ht]
    \centering
    \renewcommand{\arraystretch}{0.8}
     \begin{adjustbox}{width=0.98\textwidth}
    \begin{tabular}{clccc|ccc|ccc}
        \toprule
        \multirow{2.5}{*}{\textbf{Type}}&
        \multirow{2.5}{*}{\textbf{Method}} 
        & \multicolumn{3}{c|}{\textbf{Arithmetic + Calculator}} 
        & \multicolumn{3}{c|}{\textbf{TriviaQA + RAG}} 
        & \multicolumn{3}{c}{\textbf{Math500 + Reasoner}} \\
        \cmidrule(lr){3-5} \cmidrule(lr){6-8} \cmidrule(lr){9-11}
        && \textbf{Acc $\uparrow$} & \textbf{Tool Rate $\downarrow$} & \textbf{Utility(0.2) $\uparrow$} 
        & \textbf{Acc $\uparrow$} & \textbf{Tool Rate $\downarrow$} & \textbf{Utility(0.4) $\uparrow$}
        & \textbf{Acc $\uparrow$} & \textbf{Tool Rate $\downarrow$} & \textbf{Utility(0.6) $\uparrow$} \\
        \midrule        \rowcolor{gray!8}\multicolumn{11}{c}{\textit{Llama3.1 8B}}\\
        \midrule
        \multirow{4}{*}{Prompt-based} & Baseline (w/o tool) & 63.0 & 0.0 & 63.0 & 62.5 & 0.0 & 62.5 & 51.4 & 0.0 & 51.4 \ \\
        & Baseline (all tool) & 99.0 & 100.0 & 79.0 & 95.8 & 100.0 & 55.8 & 96.2 & 100.0 & 36.2 \ \\
        & Auto tool & 90.3 & 75.0 & 75.3 & 89.5 & 78.0 & 58.3 & 73.1 & 50.1 & 43.0 \ \\
        & ICL tool (10-shot) & 91.6 & 62.6 & 79.2 & 85.6 & 69.5 & 57.8 & 53.2 & 4.9 & 50.3 \ \\
        \midrule
        \multirow{4}{*}{Uncertainty-based} & Raw logits  & 90.7 & 54.6 & 79.8 & 74.3 & 16.9 & 67.5 & 59.0 & 9.9 & 53.1 \ \\
        & P(True)  & 90.4 & 65.1 & 77.4 & 87.4 & 59.2 & 63.7 & 84.4 & 61.6 & 47.4 \ \\
        & verb. 1S top-1  & 65.5 & 7.8 & 63.9 & 77.4 & 32.8 & 64.3 & 64.1 & 16.3 & 54.3 \ \\
        & verb. 2S top-1  & 69.1 & 16.1 & 65.9 & 74.8 & 20.9 & 66.3 & 62.0 & 16.7 & 52.0  \ \\
        & agreement(consistency) & 77.3 & 22.4 & 72.8 & 87.3 & 45.7 & 69.0 & 71.7 & 28.5 & 54.6 \ \\
        \midrule
        \multirow{4}{*}{Training-based} 
         & \textsc{Implicit-Consistency} & 80.1 & 30.9 & 73.9 & 77.0 & 25.1 & 67.0 & 84.4 & 51.6 & 53.6 \ \\
         & \textsc{Implicit-Absolute} & 96.7 & 45.2 & \textbf{87.7} & 91.1 & 42.3 & \textbf{74.2} & 93.1 & 55.5 & \textbf{59.8}   \ \\
         & \textsc{Explicit-Consistency} & 90.7 & 61.7 & 78.4 & 76.9 & 25.9 & 66.5 & 84.1 & 45.7 & 56.7 \ \\
          & \textsc{Explicit-Absolute} & 93.3 & 33.8 & 86.5 & 82.9 & 29.7 & 71.0 & 79.5 & 35.6 & 58.1  \ \\
        \midrule
        \rowcolor{gray!8}\multicolumn{11}{c}{\textit{Qwen2.5 7B}}\\
        \midrule
        \multirow{4}{*}{Prompt-based} & Baseline (w/o tool)  & 67.0 & 0.0 & 67.0 & 51.1 & 0.0 & 51.1 & 74.9 & 0.0 & 74.9 \ \\
        & Baseline (all tool) & 99.0 & 100.0 & 79.0 & 94.7 & 100.0 & 54.7  & 96.2 & 100.0 & 36.2  \ \\
        & Auto tool & 95.7 & 83.4 & 79.0 & 90.4 & 89.6 & 54.6 & 77.1 & 24.5 & 62.4 \ \\
        & ICL tool (10-shot) & 91.2 & 32.9 & 84.6 & 74.5 & 33.8 & 61.0 & 75.1 & 1.8 & 74.0 \ \\
        \midrule
        \multirow{4}{*}{Uncertainty-based} & Raw logits  & 95.1 & 47.8 & 85.5 & 86.6 & 61.7 & 61.9 & 86.9 & 34.1 & 66.4 \ \\
        & P(True)  & 94.2 & 63.4 & 81.5 & 79.1 & 53.1 & 57.9 & 86.0 & 30.7 & 67.6 \ \\
        & verb. 1S top-1  & 68.9 & 4.9 & 67.9 & 81.2 & 55.9 & 58.8 & 75.6 & 6.9 & 71.5 \ \\
        & verb. 2S top-1  & 78.9 & 22.4 & 74.4 & 79.5 & 51.5 & 58.9 & 83.9 & 20.2 & 71.8 \ \\
        & agreement(consistency) & 91.6 & 22.4 & 87.1 & 86.2 & 47.9 & 67.0 & 97.8 & 38.6 & 74.6 \ \\
        \midrule
        \multirow{4}{*}{Training-based} 
         & \textsc{Implicit-Consistency}  & 82.7 & 17.2 & 79.3 & 84.2 & 58.1 & 61.0 & 96.9 & 54.9 & 64.0  \ \\
         & \textsc{Implicit-Absolute} & 97.6 & 37.9 & 90.1 & 90.7 & 59.1 & 67.1 & 93.9 & 29.0 & 76.5 \ \\
         & \textsc{Explicit-Consistency} & 90.7 & 61.7 & 78.4 & 72.9 & 23.9 & 63.3 & 89.9 & 22.3 & 76.5 \ \\
         & \textsc{Explicit-Absolute} & 97.3 & 28.8 & \textbf{91.5} & 80.3 & 30.3 & \textbf{68.2} & 90.1 & 21.2 & \textbf{77.4}  \ \\
        \midrule
    \end{tabular}%
    \end{adjustbox}
    \caption{Performance comparison on three tool calling scenarios. The utility is the overall evaluation metric of accuracy and tool rate. A larger $\alpha$ indicates a higher cost sensitivity and a greater penalty for invoking tools.}
        
    \label{tab:main_results}
\end{table*}

\subsection{Overconfidence and Over-tool-reliance}

\begin{figure}[t]
    \centering
    \includegraphics[width=\linewidth]{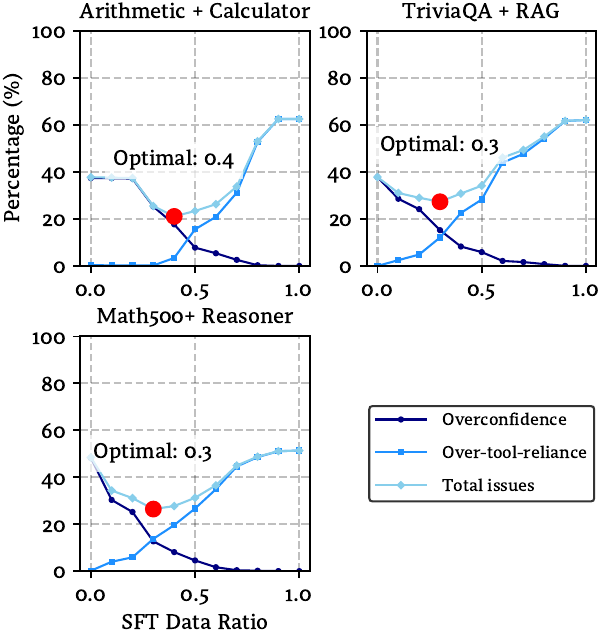}
    \caption{Trade-off between overconfidence and over-tool-reliance with different SFT data ratios.}

    \label{fig:explicit_overconfidence_tooluse}
\end{figure}

We analyze how explicit modeling affects the trade-off between overconfidence and over-tool-reliance by adjusting the SFT data ratio, which represents the proportion of training samples with tool invocation. As this ratio increases, the model's confidence estimation and reliance on external tools shift. Figure~\ref{fig:explicit_overconfidence_tooluse} shows the relationship between the SFT data ratio and the combined proportion of overconfidence and over-tool-reliance. Higher tool usage increases reliance on external tools, leading to excessive tool invocation, while overconfidence gradually decreases as the model defers more to tools. Each dataset exhibits an optimal SFT data ratio, where this combined proportion is minimized, balancing model confidence and tool dependency. This turning point in Figure~\ref{fig:explicit_overconfidence_tooluse} serves as a guideline for optimal model selection. At this ratio, the model maintains a well-calibrated knowledge boundary while minimizing unnecessary tool usage, ensuring both efficiency and accuracy.

\subsection{Inference Time}

\begin{figure}[t]
    \centering
    \includegraphics[width=\linewidth,trim=200 0 200 0,clip]{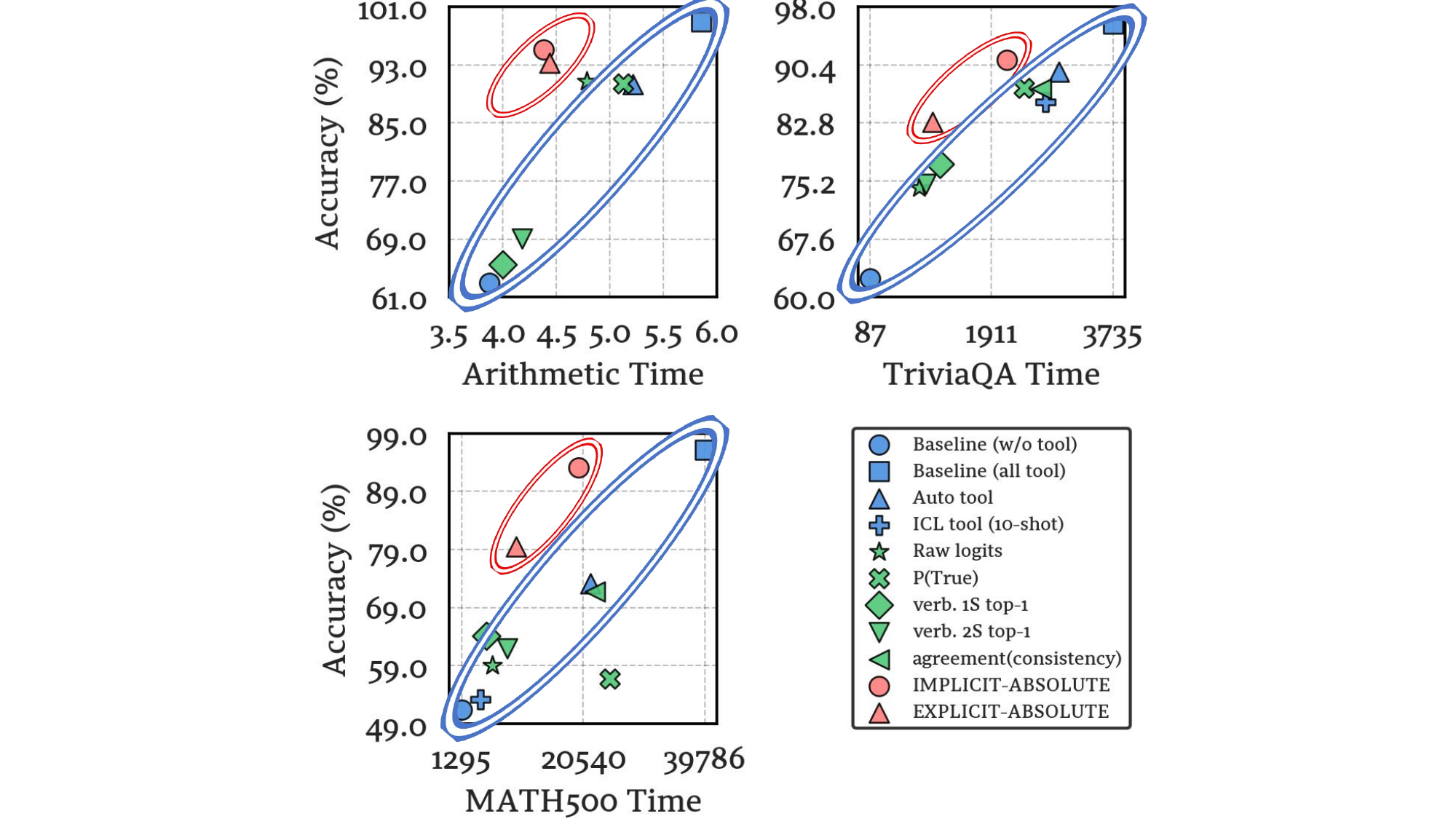}
    \caption{Performance vs. inference time (seconds).}
    \label{fig:time}
    \vspace{-5mm}
\end{figure}

Since tool invocation adds computational overhead, we assess inference cost by measuring actual execution time. Using VLLM~\cite{kwon2023efficient} on NVIDIA A800 GPUs(see Appendix~\ref{sec:inference set up} for detailed experimental setup), we compute per-sample inference time and aggregate the total runtime across the dataset. Figure~\ref{fig:time} illustrates the trade-off between inference time and performance, where methods positioned towards the upper-left corner achieve a more favorable balance. Our approach consistently demonstrates superior efficiency, attaining either higher performance at the same inference time or reduced latency while maintaining accuracy. By optimizing tool usage, our method minimizes cost without performance loss, ensuring efficient real-world deployment and making it suitable for practical applications.


\subsection{Ablation}

\subsubsection{Implict Modeling Methods}

\begin{figure*}
    \centering
    \includegraphics[width=\linewidth]{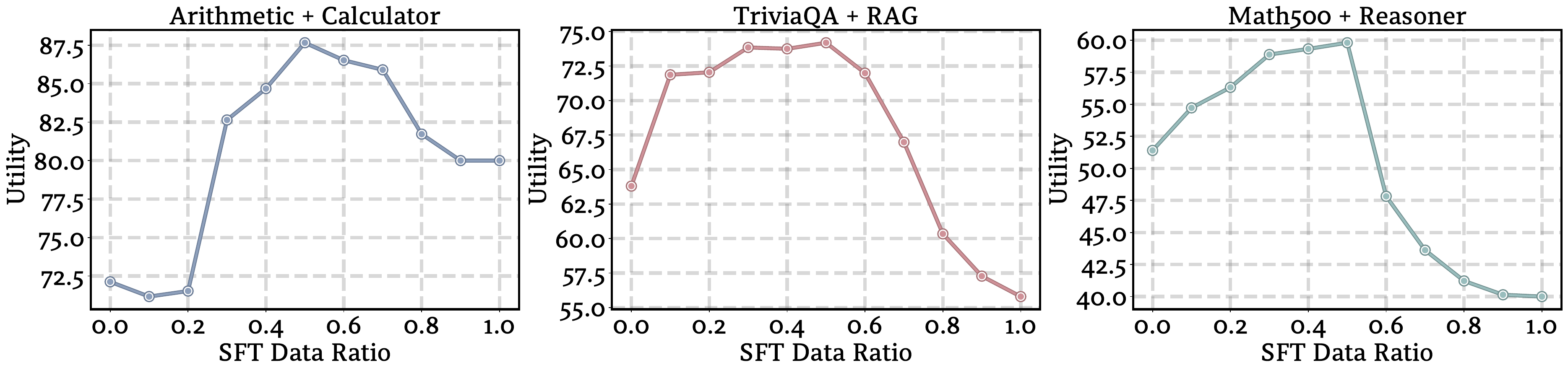}
    \caption{\textbf{Effect of SFT Data Ratio on Utility.} The ratio represents the proportion of training samples in which the model invokes a tool rather than answering directly.}
    \label{fig:ablation_sft_ratio}
\end{figure*}

To understand how implicit modeling affects our results, we perform an ablation study to see how different Supervised Fine-Tuning (SFT) data ratios impact the model's behavior. The data ratio means the percentage of training examples where the model uses a tool to get the answer instead of answering on its own. We keep the total dataset size the same but change this ratio to see how it affects the model's preference for using tools or answering directly. This helps us find the best balance based on cost.
When using a tool is cheap, a higher ratio makes the model use tools more often, which improves accuracy by using external resources. On the other hand, if tool usage is expensive, a lower ratio makes the model answer questions independently, reducing costs. The key is to find the right balance so the model efficiently decides when to use tools based on the situation.
Figure~\ref{fig:ablation_sft_ratio} shows how the data ratio affects the model's utility. At first, utility increases as the ratio goes up, reaching a peak before dropping. The best ratio is different for each dataset and depends on how much the tool costs. If tool costs are high, the optimal ratio is lower. This shows that our implicit modeling approach helps the model make smart choices based on task costs, balancing accuracy and efficiency.

\subsubsection{Explicit Modeling Methods}
\begin{figure}[t]
    \centering
    \includegraphics[width=\linewidth]{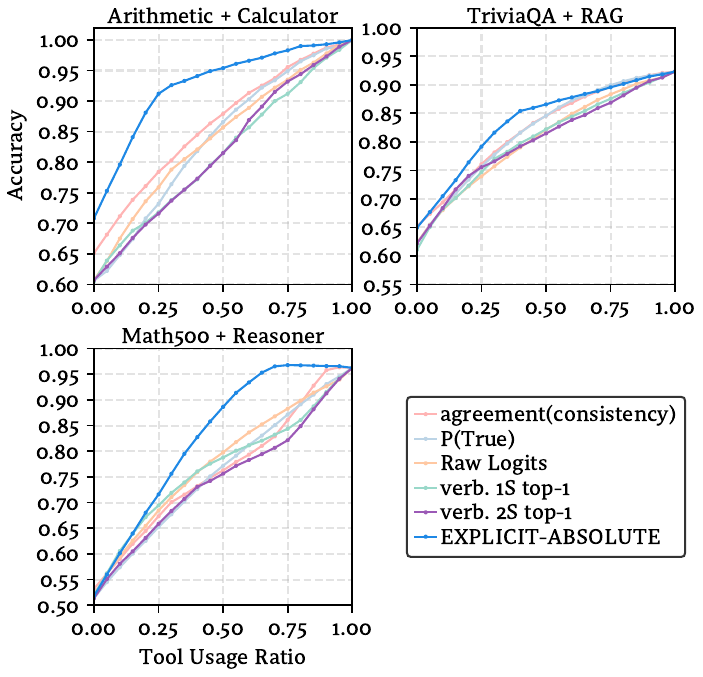}
 \caption{Comparison of tool invocation strategies: explicit modeling vs. uncertainty-based baselines.}

    \label{fig:explicit_ablation}
\end{figure}

Unlike implicit approaches, explicit modeling allows the model to directly output confidence scores alongside its predictions, enabling threshold-based decision-making for tool invocation. To further evaluate its effectiveness, we compare explicit modeling with uncertainty-based baselines, as both methods fundamentally rely on confidence estimation to determine knowledge boundaries. To ensure a fair comparison, we adjust the confidence threshold to control the tool invocation ratio, systematically varying the threshold to assess model performance at different levels of tool usage. As shown in Figure~\ref{fig:explicit_ablation} illustrates the relationship between tool invocation rate and model performance across various confidence thresholds. Explicit modeling consistently outperforms uncertainty-based baselines at all invocation ratios, demonstrating its ability to provide a more reliable estimation of knowledge boundaries. The performance gap remains stable, highlighting the robustness of explicit confidence modeling. By leveraging these confidence scores, our approach enables finer control over tool invocation, optimizing task success while reducing unnecessary computational overhead.

\subsection{Knowledge Boundary Distribution After Alignment}

\begin{figure}[t]
    \centering
    \includegraphics[width=\linewidth]{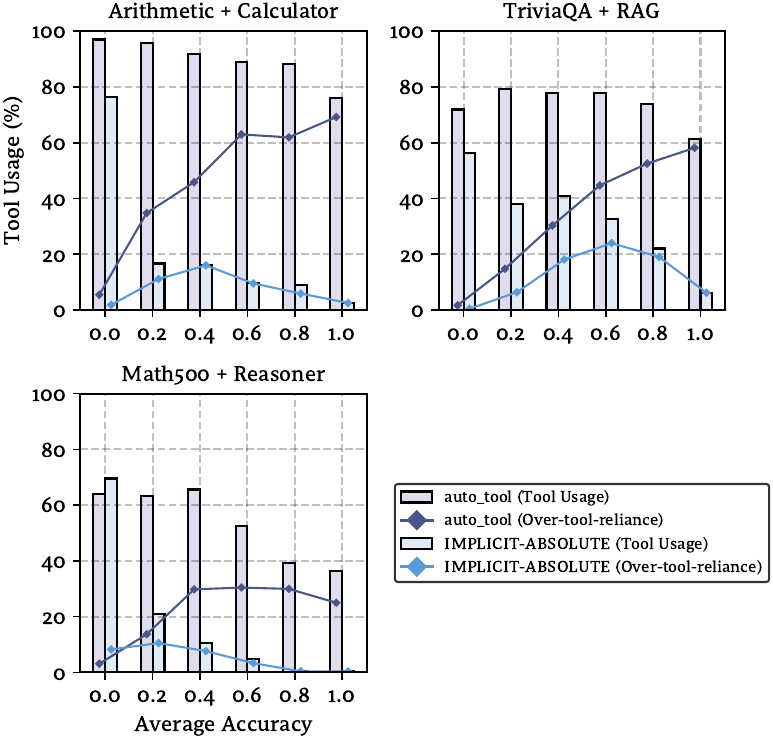}
    \caption{Comparison of tool usage and over-tool-reliance across different accuracy levels.}

\label{fig:distribution}
\end{figure}
To examine whether the model learns about knowledge boundary, we compare our method with  \texttt{auto\_tool} in terms of tool invocation distribution.
Figure~\ref{fig:distribution} presents tool usage across different accuracy levels. Higher accuracy reflects a better understanding of the problem. An ideal model should rely on tools for challenging cases while minimizing tool use for confidently answered questions. However, \texttt{auto\_tool} exhibits a nearly uniform tool invocation pattern, suggesting it lacks awareness of its knowledge boundaries. In contrast, our method shows a gradual decline in tool usage as accuracy increases, indicating adaptive tool invocation based on knowledge confidence.
We also analyze over-tool-reliance, where the model uses tools unnecessarily despite being capable of answering correctly. Figure~\ref{fig:distribution} shows that the baseline exhibits increasing over-tool-reliance with accuracy, leading to unnecessary computational overhead. Conversely, our method reduces over-tool-reliance, enabling more intelligent invocations.


\section{Conclusion}

In this work, we introduced a novel approach to improve LLMs' decision-making regarding when and how to use external tools. By incorporating the concept of an "uncertain region" and probabilistic knowledge boundary estimation, our framework enables more informed and efficient tool usage. Through extensive experiments, we demonstrated that our approach reduces unnecessary tool calls, improving performance and cost-effectiveness. By combining implicit and explicit modeling techniques, we provide the model with greater flexibility in real-time decisions. Our work advances LLMs' tool intelligence, ensuring more judicious and efficient tool invocation. Future work can explore further refinements and broader applications.

\section*{Limitations}
This work primarily proposes an alignment framework for efficient tool invocation, evaluated through experiments on three datasets. On the one hand, the number of tools used in these experiments is limited, with a selection of three representative tools: a mathematical calculator, a search engine, and an external large model. This choice is motivated by the fact that most tools possess highly specific knowledge. For example, tools that retrieve weather information for a particular day contain knowledge that does not overlap with that of the model, requiring the model to invoke the tool to complete the task. On the other hand, different models and knowledge sources can also be framed as tools, meaning that the discussion in this work on modeling knowledge boundaries remains highly valuable. In addition, the experiments in this work were conducted on only two open-source models, as obtaining baseline data for closed-source models presents significant challenges. For instance, methods such as uncertainty estimation often require access to specific token logits, which are difficult to obtain for proprietary models. This limitation affects the generalizability of the experimental results, as the performance of closed-source models may differ in ways that cannot be captured without direct access to their internals.

\bibliography{custom}

\appendix

\label{sec:appendix}

\section{Uncertainty Estimation Methods}
\label{sec:uncertainty}

This section provides a comprehensive overview of the uncertainty estimation techniques employed in our study. These methods aim to quantify model confidence in its predictions, helping regulate tool invocation and decision-making.

\paragraph{Raw Logits.}  
This approach estimates confidence using the model's logit values, specifically by computing the exponential of the average log probability of the generated tokens. This metric is mathematically equivalent to the reciprocal of perplexity, where lower perplexity indicates higher confidence, effectively capturing how certain the model is in its prediction.

\paragraph{Agreement (Consistency-based).}  
In this method, confidence is determined by measuring the proportion of generated responses that align with the most frequently predicted answer. A higher agreement percentage suggests greater internal consistency in the model's responses, thereby indicating a stronger level of confidence in its generated output.

\paragraph{P(True).}  
This method involves prompting the model to explicitly assess the correctness of its own response. The confidence score is derived from the normalized probability assigned to the ‘True’ token, reflecting the model's self-evaluated likelihood that its answer is correct.

\paragraph{Verbalized Confidence: 1-Stage Top-k (Verb. 1S Top-k).}  
In this one-stage approach, the model generates the top \( k \) candidate answers along with their respective probabilities in a single pass. The highest-ranked answer and its assigned probability serve as an indicator of confidence, offering a direct estimation of the model’s certainty in its response.

\paragraph{Verbalized Confidence: 2-Stage Top-k (Verb. 2S Top-k).}  
Unlike the single-stage method, this two-stage approach first prompts the model to generate multiple candidate answers and then separately assigns probabilities to each of them in a second inference step. The final confidence score is computed based on these probabilities, allowing for a refined estimation that accounts for potential self-correction.

These uncertainty estimation techniques play a crucial role in calibrating tool invocation decisions, ensuring that external tools are utilized effectively based on the model’s confidence in its own predictions.
To optimize utility, we sort all confidence scores across responses and use each unique score as a potential threshold, systematically evaluating its impact on tool invocation. 

\section{Experimental Setup}

\label{sec:setup}
\paragraph{Arithmetic Computation.}  
For arithmetic tasks, we use a dataset consisting of  10,000 training samples and  1,000 test samples. To ensure the quality of generated arithmetic expressions, we filter out any syntactically incorrect or malformed expressions that do not conform to standard arithmetic formats. Symbolic computation is performed using the  SymPy library, which provides a robust framework for symbolic mathematics and equation evaluation.

\paragraph{Knowledge-based QA (TriviaQA).}  
For knowledge-based question answering, we randomly select  10,000 training instances from the full TriviaQA training set. The retrieval system is employed  only during inference and does not participate in training. During training, the model is only exposed to the tool invocation format, but actual retrieval is not performed. We follow the Pyserini setup for TriviaQA and utilize a  sparse retriever to retrieve the top  100 highest-scoring passages. To improve retrieval accuracy, we further filter passages that contain the correct answer and refine the selection using  ChatGPT, eliminating irrelevant noisy passages. This ensures that the retrieved information is reliable, preventing erroneous tool invocation from negatively impacting final performance.

\paragraph{Complex Reasoning (MATH).}  
For mathematical problem-solving, we process the  MATH dataset following its original settings. We employ  DeepSeek-R1 (671B) as the external reasoning model, deploying it locally using  VLLM on a cluster of  32 NVIDIA A800 GPU. The model operates in a  zero-shot setting. To mitigate excessive inference latency, we instruct the model to generate concise responses while maintaining reasoning completeness. Despite this constraint, DeepSeek-R1 still significantly surpasses our primary models in response time.

\section{Inference Time Experimental Setup}
\label{sec:inference set up}
For inference time evaluation, we employ the  VLLM framework and conduct experiments on two NVIDIA A800 GPUs. To obtain a precise measurement of raw inference latency, we process input samples sequentially, without applying any parallelization techniques such as batching. We measure only the pure inference time, excluding any overhead from data loading. All other parameters remain at their default settings, and the model is loaded in  bfloat16 format to optimize memory usage while preserving numerical precision.

\section{Prompts Used in Experiments}
\label{sec:Prompts}
\subsection{Prompts Used in Different Prompt-based Methods}
\label{sec:Prompt-based}

The prompts used for different datasets are presented in the following sections. 
Table~\ref{tab:PromptsMATH} shows the prompts for the \textbf{MATH dataset}, 
Table~\ref{tab:PromptsArithmetic} contains the prompts for the \textbf{Arithmetic dataset}, 
and Table~\ref{tab:PromptsTriviaQA} presents the prompts for the \textbf{TriviaQA dataset}.

\subsubsection{Prompts for MATH Dataset}
\label{sec:Prompt-MATH}
Table~\ref{tab:PromptsMATH} lists the prompts used for different methods when evaluating the MATH dataset.

\subsubsection{Prompts for Arithmetic Dataset}
\label{sec:Prompt-Arithmetic}
Table~\ref{tab:PromptsArithmetic} lists the prompts used for different methods when evaluating the Arithmetic dataset.

\subsubsection{Prompts for TriviaQA Dataset}
\label{sec:Prompt-TriviaQA}
Table~\ref{tab:PromptsTriviaQA} lists the prompts used for different methods when evaluating the TriviaQA dataset.

\subsection{Prompts Used in Different Uncertainty-based Methods}
\label{sec:Uncertainty-based}
The prompts are shown in Table~\ref{tab:Uncertainty-based}.

\begin{table*}[ht!]
\small
    \centering
          \begin{tabular}{m{\linewidth}<{\raggedright}}
        \toprule
        \rowcolor[gray]{0.95} 
        \textbf{Baseline (w/o tool) - MATH} \\
        \midrule

            Given the following problem, break it down into steps and reason through each part before arriving at a final conclusion. 
            Your final answer MUST be enclosed in \textbackslash boxed\{\}. \\
            Problem: \{question\} \\

        \midrule
        \rowcolor[gray]{0.95} 
        \textbf{Baseline (all tool) - MATH} \\
        \midrule

            Given the following problem, break it down into steps and reason through each part before arriving at a final conclusion. 
            Your final answer MUST be enclosed in \textbackslash boxed\{\}. \\
            Problem: \{question\} \\

        \midrule
        \rowcolor[gray]{0.95} 
        \textbf{Auto Tool - MATH} \\
        \midrule

            Given the following problem. If you can solve it directly with confidence, your final answer must be in \textbackslash boxed\{\} format. 
            If you cannot solve it directly, call the tool immediately without reasoning, using this format: \\
            
            \{\{ \\
            \quad "tool\_name": "math\_solver" \\
            \}\} \\
            
            Problem: \{question\} \\

        \midrule
        \rowcolor[gray]{0.95} 
        \textbf{ICL Tool (10-shot) - MATH} \\
        \midrule

            Given the following problem. If you can solve it directly with confidence, your final answer must be in \textbackslash boxed\{\} format. 
            If you cannot solve it directly, call the tool immediately without reasoning, using this format: \\
            
            \{\{ \\
            \quad "tool\_name": "math\_solver" \\
            \}\} \\
            
            Examples: \{example\} \\
            
            Problem: \{question\} \\

        \bottomrule
    \end{tabular}
    
    \caption{Prompts Used in Different Methods for MATH Dataset.}
    \label{tab:PromptsMATH}
\end{table*}

\begin{table*}[ht!]
\small
    \centering
          \begin{tabular}{m{\linewidth}<{\raggedright}}
        \toprule
        \rowcolor[gray]{0.95} 
        \textbf{Baseline (w/o tool) - Arithmetic} \\
        \midrule

            Given the following problem, provide the final answer directly. \\
            Problem: \{question\} \\
            Your response should only be "The final answer is [answer]" where [answer] is the response to the problem. \\

        \midrule
        \rowcolor[gray]{0.95} 
        \textbf{Baseline (all tool) - Arithmetic} \\
        \midrule

            Use a calculator to solve the question. Format your output as a JSON object in the following structure: \\
            
            \{\{ \\
            \quad "calculator": "<expression>" \\
            \}\} \\
            
            Problem: \{question\} \\

        \midrule
        \rowcolor[gray]{0.95} 
        \textbf{Auto Tool - Arithmetic} \\
        \midrule

            If you are confident in your answer, output the final answer directly. 
            If unsure, use the calculator tool and respond with a JSON object formatted as: \\
            \\
            \{\{ \\
            \quad "tool\_name": "calculator" \\
            \}\} \\
            Problem: \{question\} \\

        \midrule
        \rowcolor[gray]{0.95} 
        \textbf{ICL Tool (10-shot) - Arithmetic} \\
        \midrule

            If you are confident in your answer, output the final answer directly. 
            If unsure, use the calculator tool and respond with a JSON object formatted as: \\
            \\
            \{\{ \\
            \quad "tool\_name": "calculator" \\
            \}\} \\
            \\
            Examples: \{example\} \\
            Problem: \{question\} \\

        \bottomrule
    \end{tabular}
    
    \caption{Prompts Used in Different Methods for Arithmetic Dataset.}
    \label{tab:PromptsArithmetic}
\end{table*}

\begin{table*}[ht!]
\small
    \centering
          \begin{tabular}{m{\linewidth}<{\raggedright}}
        \toprule
        \rowcolor[gray]{0.95} 
        \textbf{Baseline (w/o tool) - TriviaQA} \\
        \midrule

            Answer the following question. 
            Your response should only be "The final answer is [answer]" where [answer] is the response to the problem. \\
            
            Problem: \{question\} \\

        \midrule
        \rowcolor[gray]{0.95} 
        \textbf{Baseline (all tool) - TriviaQA} \\
        \midrule

            \{documents\} \\
            
            Based on the information in this document, answer the following question accurately. \\
            
            Problem: \{question\} \\

        \midrule
        \rowcolor[gray]{0.95} 
        \textbf{Auto Tool - TriviaQA} \\
        \midrule

            Answer the following question directly if you are confident in your knowledge. 
            If you are uncertain or need to retrieve information, respond with a JSON object in the following format: \\
            
            \{\{ \\
            \quad "tool\_name": "search\_info" \\
            \}\} \\
            
            Problem: \{question\} \\

        \midrule
        \rowcolor[gray]{0.95} 
        \textbf{ICL Tool (10-shot) - TriviaQA} \\
        \midrule

            Answer the following question directly if you are confident in your knowledge. 
            If you are uncertain or need to retrieve information, respond with a JSON object in the following format: \\
            
            \{\{ \\
            \quad "tool\_name": "search\_info" \\
            \}\} \\
            
            Examples: \{example\} \\
            
            Problem: \{question\} \\

        \bottomrule
    \end{tabular}
    
    \caption{Prompts Used in Different Methods for TriviaQA Dataset.}
    \label{tab:PromptsTriviaQA}
\end{table*}

\begin{table*}[ht!]
\small
    \centering
        \begin{tabular}{m{\linewidth}<{\raggedright}}
        \toprule
        \rowcolor[gray]{0.95} 
        \textbf{Logits-based  Prompt} \\
        \midrule

            You are a helpful assistant. \\
            
            Answer the following question as accurately as possible. \\
            Question: \{question\} \\

        \midrule
        \rowcolor[gray]{0.95} 
        \textbf{P(true) Prompt} \\
        \midrule

            You are a helpful assistant. You should judge whether the answer to the given question is True or False. 
            Please only reply with a simple word "True" or "False". \\
            Answer the following questions as accurately as possible. \\
            Question: \{question\} \\
            Answer: \{answer\} \\
            Is the above answer correct? (True / False) \\

        \midrule
        \rowcolor[gray]{0.95} 
        \textbf{Consistency Prompt} \\
        \midrule

            You are a helpful assistant. \\
            
            Answer the following question as accurately as possible. Provide ONLY the direct answer without any explanation. \\
            Question: \{question\} \\

        \midrule
        \rowcolor[gray]{0.95} 
        \textbf{Verb. 1S top1 Prompt} \\
        \midrule

            You are a helpful assistant, and you are always completely honest and DIRECT in your responses. \\
            
            Provide a brief, concise answer along with an explicit confidence percentage (0-100\%) about the correctness of your response. \\
            Question: \{question\} \\

        \midrule
        \rowcolor[gray]{0.95} 
        \textbf{Verb. 2S top1 Prompt} \\
        \midrule

            You are a helpful assistant, always completely honest and direct in your responses. 
            You are also transparent about your confidence level and can honestly share how certain you are about the answer. \\
            
            Question: \{question\} \\
            Answer: \{previous\_answer\} \\
            How confident are you in the above answer (0-100\%)? \\

        \bottomrule
    \end{tabular}
    \caption{Prompts Used in Different Belief Estimation Methods.}
    \label{tab:Uncertainty-based}
\end{table*}


\end{document}